\documentclass[journal]{IEEEtran}

\usepackage{cite}
\usepackage{amsmath,amssymb,amsfonts}
\usepackage{algorithmic}
\usepackage{graphicx}
\usepackage{textcomp}
\usepackage{xcolor}
\usepackage{url}
\usepackage{booktabs}
\usepackage{multirow}
\usepackage{array}

\usepackage[hidelinks]{hyperref}

\def\BibTeX{{\rm B\kern-.05em{\sc i\kern-.025em b}\kern-.08em
    T\kern-.1667em\lower.7ex\hbox{E}\kern-.125emX}}

\newcommand{\sloc}{S_{\text{loc}}}
\newcommand{\scls}{S_{\text{cls}}}
\newcommand{\sscr}{S_{\text{screen}}}
\newcommand{\sone}{S_{\text{Track1}}}

\begin{document}

\title{Cross-Scenario Industrial Defect Detection via Domain-Robust Training and Metric-Aligned Selection}

\author{Dikai Li, Weijia Wang*, and Yuan-Gen Wang*\\
School of Artificial Intelligence, Guangzhou University, Guangzhou 510006, China\\
dikaili@e.gzhu.edu.cn, wwj@gzhu.edu.cn, wangyg@gzhu.edu.cn
\thanks{*Corresponding author. This work was partly supported by the National Natural Science Foundation of China under Grant No. 62502108.}
\thanks{This paper has been accepted for presentation in IEEE ICME 2026.}}


\maketitle

\begin{abstract}
Automated defect detection of industrial boards must remain highly reliable under cross-scenario variations in lighting, surface contaminants, and production line conditions. This makes cross-scenario defect detection a domain generalization problem, wherein models are required to detect, localize, and classify defects across unseen scenarios while minimizing false alarms on clean boards. For the IDA 2026 Track 1 challenge, we propose a cross-scenario defect detection method that includes four complementary modules: cross-scenario robust training, small-defect-aware learning, score calibration and inference threshold selection, and metric-aligned checkpoint selection. These four modules respectively introduce cross-scenario robust normalization and DSU-style statistic perturbation with progressive random convolution, NWD and parameter-free spatial attention, the VFL learning objective, and a custom SWAD variant driven by the Track 1-oriented proxy. Experimental results demonstrate that our model achieves a Track 1 score of 0.564 with a screening score of 0.696 on the test dataset on a single RTX 4090 GPU. The code is available at https://github.com/GZHU-DVL/Dikai.

\end{abstract}


\section{Introduction}

Automated visual inspection algorithms of printed-circuit board assemblies and related industrial board products must remain reliable under various conditions, such as lighting, surface contamination, and production line configurations. In such settings, training images are typically collected from limited source scenarios, whereas deployment may involve samples with unseen conditions. This makes industrial defect analysis (IDA) a domain-generalization problem, rather than a simple defect detection task within a single domain. A practical inspection system must not only locate defects accurately on defective boards, but also screen clean boards correctly to reduce false alarms. The IDA~2026 Track~1 benchmark~\cite{ida2026challenge} turns this requirement into a composite metric $\sone$ that aggregates scores of localization $\sloc$, classification $\scls$, and screening $\sscr$ (i.e., the ability to correctly accept clean images and reject defective ones). Among these, $\sscr$ has the largest weight and is the most sensitive metric to domain shift between the training data distribution and the test set.

A widespread workflow in such challenges is to train multiple model variants, select checkpoints according to the local validation metric, and submit the best-performing one. However, for cross-scenario industrial inspection, this procedure can be unreliable when the selection criterion is dominated by defect localization accuracy. During deployment, a checkpoint with strong defect detection performance may produce excessive false alarms on defect-free boards after the appearance distribution changes. Therefore, checkpoint selection and score calibration should be treated as critical components of cross-scenario generalization, rather than as regular post-training choices. 

\begin{figure*}[t]
\centering
\includegraphics[width=\textwidth]{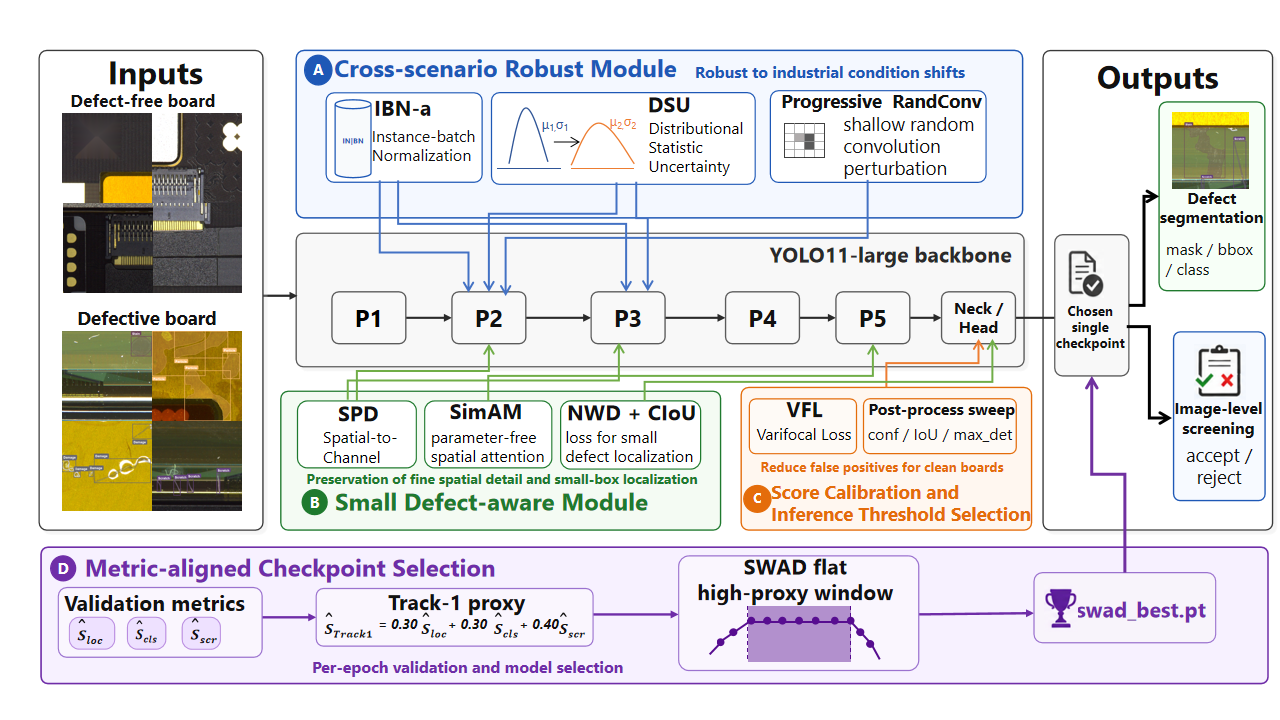}
\caption{Overview of the proposed method for cross-scenario industrial defect detection.}
\label{fig:pipeline}
\end{figure*}

Motivated by this observation, we design our framework with four
components. First, the Cross-Scenario Robust Module improves domain robustness through IBN-a and MixRand. IBN-a splits shallow feature channels into instance-normalized and batch-normalized parts, while MixRand combines DSU and PRC to perturb feature statistics and local textures during training. Second, the Small-Defect-Aware Module preserves fine-grained spatial cues and strengthens localization signals for defects that occupy only a small image region. Third, the Score Calibration and Inference Threshold Selection calibrates confidence scores and selects deployment thresholds to reduce false alarms on clean boards. Finally, the Metric-Aligned Checkpoint Selection replaces localization-dominated checkpoint ranking with a Track-1-oriented proxy and SWAD-based averaging, so that the selected checkpoint better matches the official screening-aware objective.

This paper formulates IDA~Track~1 as a cross-scenario domain-generalization problem for industrial defect detection. The screening-focused evaluation highlights the limitations of standard mAP-driven selection, and motivates a framework that improves appearance robustness, enhances small-defect detection, calibrates scores and inference thresholds, and aligns checkpoint selection with the objective metric. In summary, our contributions are summarized as follows:
\begin{itemize}
\item We identify cross-scenario industrial defect detection on IDA~Track~1 as a deployment-oriented domain-generalization problem, and show that mAP-driven checkpoint selection can be misaligned with cross-scenario screening reliability.

\item We present a practical single-model framework that combines appearance-robust training, small-defect preservation, screening-aware calibration, and metric-aligned checkpoint averaging for robust industrial-board inspection under domain shift.

\item We provide targeted ablations showing that metric-aligned SWAD selection is the clearest measured contributor under a unified local protocol, while composite feature perturbation mainly acts as a stability regularizer.

\item With no ensemble, external data, pseudo-labels, or test-time augmentation, our best queried official challenge test feedback reaches $\sone{=}0.564$ with $\sscr{=}0.696$ on a single RTX~4090 GPU.
\end{itemize}

\section{Related Work}
\textbf{Industrial anomaly and defect benchmarks.}
Industrial inspection benchmarks have provided important testbeds for visual
anomaly detection, defect localization, and defect detection. MVTec AD
\cite{bergmann2019mvtec} introduced a widely used real-world benchmark for
unsupervised anomaly detection across diverse object and texture categories,
while VisA \cite{zou2022spot} further expanded industrial anomaly detection
toward more complex visual categories and detection-oriented evaluation.
Compared with these benchmarks, IDA 2026 Track 1 emphasizes cross-scenario
industrial board inspection, where models are evaluated under unseen
production conditions and image-level screening reliability is explicitly
included in the final metric. This setting makes the task closer to
deployment-oriented defect detection than to single-domain anomaly
detection.

\textbf{Domain generalization for cross-scenario detection.}
Domain generalization aims to improve model transfer to unseen domains without
using target-domain training data \cite{zhou2023dgsurvey}. For visual
recognition and detection, a common strategy is to reduce the model's
dependence on domain-specific appearance statistics. MixStyle
\cite{zhou2021mixstyle} mixed feature statistics across samples to regularize
style-dependent representations, while DSU \cite{li2022uncertainty} modelled
uncertainty in feature statistics to improve out-of-distribution
generalization. RandConv \cite{xu2021randconv} perturbed local texture patterns
through random convolutions, encouraging models to rely less on superficial
texture cues. IBN \cite{pan2018ibn} combined Instance Normalization and Batch
Normalization to balance appearance invariance and discriminative feature
learning. In addition to architectural and training-time regularization,
SWAD \cite{cha2021swad} averaged weights within flat regions of the training
trajectory and has been shown to improve domain-generalized solutions. These
studies motivate domain-robust training and checkpoint averaging for
cross-scenario industrial inspection, where appearance statistics may change
between training and the challenge test scenarios, whose labels are not publicly available.

\textbf{Tiny defect detection, attention, and score calibration.}
Industrial defects often occupy only a small fraction of an image, making
their localization sensitive to downsampling, feature aggregation, and sparse
IoU gradients. NWD \cite{wang2021nwd} modelled bounding boxes as Gaussian
distributions and used a normalized Wasserstein distance, providing a smoother
localization signal for tiny objects than IoU-only regression. SPD
\cite{sunkara2023spdconv} preserved fine spatial details by rearranging local
spatial neighborhoods into channel representations before projection, which
is useful when small visual cues may be lost during strided downsampling.
Attention mechanisms can further strengthen discriminative local responses
when defect cues are subtle and visually close to background textures. SimAM
\cite{yang2021simam} estimated neuron importance from activation statistics
and generates parameter-free spatial attention, making it lightweight for
insertion into detection backbones. Beyond localization and feature
refinement, screening-oriented inspection also requires reliable confidence
estimates. Varifocal Loss \cite{zhang2021varifocalnet} improved
classification-score calibration by emphasizing high-quality positive samples
and reducing unreliable high-confidence predictions. These lines of work
provide useful tools for cross-scenario industrial inspection, but the IDA
Track 1 setting further requires aligning training, inference thresholds, and
checkpoint selection with a screening-aware evaluation objective.

\section{Proposed Method}
Fig.~\ref{fig:pipeline} shows an overview of our proposed method. We start
from a YOLOv11-large segmentation backbone~\cite{ultralytics2024yolo11}
(initialized from public pretrained weights) and introduce four complementary modules to improve cross-scenario defect detection robustness: 1) Cross-Scenario Robust Module aiming to improve appearance robustness against lighting, fixture, and surface-style shifts through persistent normalization and stochastic feature perturbation; 2) Small-Defect-Aware Module aiming to effectively recognize small defects that are easily lost during downsampling; 3) Score Calibration and Inference Threshold Selection aiming to improve confidence calibration and image-level screening reliability; and 4) Metric-Aligned Checkpoint Selection aiming to align checkpoint selection with the final deployment objective through the IDA Track~1-oriented proxy and SWAD-based averaging. As a result, these four modules cooperate with each other, improving both defect localization accuracy and robust inspection behavior under cross-scenario shift. Note that our detection and segmentation heads are consistent with YOLOv11. The stochastic perturbation behavior is active only during training. At inference, we use a single exported checkpoint and the same rectangular-box export path for all variants, with no ensemble or test-time augmentation according to the competition requirements.

\subsection{Cross-Scenario Robust Module}
\label{sec:mixrand}
To obtain the generalization of defect detection on various domains, we construct a Cross-Scenario Robust Module, which includes our designed MixRand and the IBN-a~\cite{pan2018ibn} components. Our MixRand introduces a Progressively Random Convolution (PRC) strategy and DSU~\cite{li2022uncertainty}, aiming to capture the defect's intrinsic structure. The IBN-a is applied with a ratio of 0.6 in the stem and stage 1 and remains active at both training and inference stages, providing a standing domain-invariance prior independent of the training schedule.

Specifically, our MixRand constructs a combo-strength attenuation rule to combine global DSU and local PRC. DSU perturbs per-channel feature statistics by sampling new mean and variance from a learned uncertainty distribution and re-normalizing the activations
(2 backbone stages, $p=0.20$, $\alpha=0.20$). PRC convolves the input (or a shallow feature map) with a random depthwise kernel of size $\{3,5\}$ at probability and strength that increase linearly with the training schedule
($p_{\text{start}}{=}0.12$, $p_{\text{end}}{=}0.30$,
$\sigma_{\text{start}}{=}0$, $\sigma_{\text{end}}{=}0.10$, with tail
decay during the final \texttt{close\_mosaic} window). Our combo-strength attenuation strategy introduce an automatic attenuation rule to reduce the DSU
 and PRC's strengths
($\text{DSU}\ p,\alpha$: $0.30{\to}0.20$;
$\text{PRC}\ p$: $0.20{\to}0.12$;
$p_{\text{end}}$: $0.45{\to}0.30$;
$\sigma_{\text{end}}$: $0.15{\to}0.10$), enabling the DSU and PRC
strengths to stay within a stable regularization range. We evaluate
this design in Sec.~\ref{sec:abl-mixrand}.

\subsection{Small-Defect-Aware Module}
\label{sec:plugins}
Industrial board defects often occupy only a very small fraction of an image. Existing detection heads tend to lose them because of downsampling and IoU gradients that are sparse at small scale. Our designed Small-Defect-Aware Module can preserve them effectively without modifying the detection head.

Specifically, our module employs SPD adapters~\cite{sunkara2023spdconv} (3 stride stages, scale $=2$, learnable $\alpha_{\text{init}}{=}0.10$ ranged at $[0,0.5]$) to reshape spatial neighborhoods into channel depth and project them back into the embedding domain via a $1{\times}1$
convolution and the strided
layers, preserving fine-grained spatial detail. To capture the salient difference between small defects and their background, we exploit SimAM~\cite{yang2021simam} in the two deep stages with $\lambda{=}10^{-4}$ to implement parameter-free spatial attention. Besides, we further combine NWD~\cite{wang2021nwd} with CIoU~\cite{zheng2020diou} to provide a dense, Gaussian-Wasserstein
localization signal when IoU saturates for small boxes. We set the NWD's weight and normalization parameter to $w{=}0.35$ and $C{=}12.8$, respectively.

\subsection{Score Calibration and Inference Threshold Selection}
\label{sec:scorecal}
We observe that the BCE loss produces a confidence distribution that is
well-calibrated for mAP50-95 but spills probability mass onto benign
textures, which directly penalizes the screening metric $\sscr$. Therefore, we employ
VFL~\cite{zhang2021varifocalnet} (with $\alpha{=}0.75$ and $\gamma{=}2.0$ in our experiments) to replace the BCE loss for score calibration and defect classification, reducing the weights of easy negatives and suppressing high-confidence false alarms on clean boards.

Building upon the above-calibrated classifier, we perform an efficient inference threshold screening postprocess at the deployment stage, aiming to achieve an optimal threshold combination. To be specific, for each candidate checkpoint we first scan the grid of the inference-time settings (confidence threshold $conf$, IoU $iou$, and maximum detection boxes $max\_det$) with the
official Track~1 evaluator. Then, we select the setting that maximizes the
composite $S_\text{Track1}$ on the local split, and use it for challenge test queries.
This setting used by the queried long-schedule checkpoint is that
$conf=0.08$, $iou=0.30$, and $max\_det=50$.

\subsection{Metric-Aligned Checkpoint Selection}
\label{sec:swad}
Ultralytics fitness \cite{ultralytics2024yolo11} used to drive checkpoint selection is dominated by mAP50-95 and correlates more strongly with defect localization than with
screening efficiency $S_{Screen}$. To efficiently select the model with good generalizability, we define an approximate detection score $\hat{S}_\text{Track1}$ as an alternative of the official detection score:
\begin{equation}
\hat{S}_\text{Track1} =0.30\,\hat{S}_\text{loc} + 0.30\,\hat{S}_\text{cls} + 0.40\,\hat{S}_\text{screen},
\label{eq:proxy}
\end{equation}
where $\hat{S}_\text{loc}$, $\hat{S}_\text{cls}$, and $\hat{S}_\text{screen}$ are the approximation of the official $S_\text{loc}$, $S_\text{cls}$, and $S_\text{screen}$, respectively. Let $p_B$, $r_B$, $\mathrm{mAP}^{B}_{50}$,
and $\mathrm{mAP}^{B}_{50:95}$ denote the box precision, recall, mAP50, and
mAP50-95, and let $p_M$, $r_M$, $\mathrm{mAP}^{M}_{50}$, and 
$\mathrm{mAP}^{M}_{50:95}$ denote the corresponding mask metrics. We
first compute the average precision and recall across the box and mask branches as follows.
\begin{equation}
\bar p = \tfrac{1}{2}(p_B + p_M), \qquad
\bar r = \tfrac{1}{2}(r_B + r_M),
\end{equation}
where $\bar p$ and $\bar r$ denote the branch-averaged precision
and recall, respectively. This averaging is a
metric-level heuristic method to balance the two prediction branches, which is distinguished from existing instance-level logical $OR$ fusion between box and mask outputs. We
then define
\begin{align}
\hat{S}_{\mathrm{loc}} &= 0.65\,\mathrm{mAP}^{M}_{50:95} + 0.35\,\mathrm{mAP}^{B}_{50:95}, \\
\hat{S}_{\mathrm{cls}} &= 0.50\,\mathrm{mAP}^{M}_{50} + 0.50\,\mathrm{mAP}^{B}_{50}, \\
\hat{S}_{\mathrm{screen}} &= 0.55\,\bar r + 0.45\,\bar p.
\end{align}
From Eqs. (3)-(5), we can observe that $\hat{S}_{\mathrm{loc}}$ emphasizes strict localization
quality through mask-dominant mAP50-95, while
$\hat{S}_{\mathrm{cls}}$ uses the looser mAP50 criterion to capture
coarse category-correct detection without over-penalizing small boundary
misalignment. From Eq. (5), the slightly larger
weight on recall reflects the higher practical cost of false negatives
than false alarms in industrial defect detection.

In the implementation, this approximation replaces the default trainer
fitness for epoch ranking and best-checkpoint selection, and is further
used to locate a stable SWAD \cite{cha2021swad} window during training. A custom SWAD driver then identifies a flat region around the smoothed peak of $\hat{S}_{\text{Track1}}$ (window $w{=}5$,
tolerance ratio $0.20$, $\geq 2$ converged epochs) and averages the
checkpoints inside this flat region into \texttt{swad\_best.pt}.
The weight of $0.40$ on $\hat{S}_{\mathrm{screen}}$ reflects the
screening-heavy structure of the Track~1 objective. The targeted
ablation in Sec.~\ref{sec:abl-modules} shows that removing SWAD
checkpoint averaging gives the largest measured drop in our current
local protocol.

\section{Experiments}

\subsection{Settings}
We train only on the officially released IDA Track~1 data,
without external data samples, pseudo-labels, model ensembles, or test-time
augmentation. All images are processed at their native resolution of
$512{\times}512$ pixels, consistent with the released dataset format.
We optimize the model using AdamW with an initial learning rate of
$3{\times}10^{-4}$, and the classification loss gain is set to $2.6$. We
adopt conservative industrial-inspection augmentation strategies, including HSV color jitter
(where \texttt{hsv\_h}${=}0.01$, \texttt{hsv\_s}${=}0.18$,
\texttt{hsv\_v}${=}0.12$), horizontal flipping (where \texttt{translate}${=}0.03$, \texttt{scale}${=}0.10$), and copy-paste with
probability $p{=}0.10$.

The queried long-schedule checkpoint is trained for 2000 epochs with
an early-stopping patience of 300 and
$\texttt{close\_mosaic}{=}120$. All targeted ablation runs use a
unified 600-epoch training budget, with an early-stopping patience of
120 and $\texttt{close\_mosaic}{=}30$. All local results are computed
under the same official-style export and scoring pipeline to ensure
reproducibility. Specifically, instance predictions are exported as
axis-aligned rectangular bounding boxes and scored using the official
Track~1 evaluation script. Unless otherwise stated, all inference is
performed using \texttt{swad\_best.pt} with $\texttt{conf}{=}0.08$,
$\texttt{IoU}{=}0.30$, and $\texttt{max\_det}{=}50$ across all
experiments.

Following the official protocol, we report $\sloc$ (localization), $\scls$
(classification), $\sscr$ (screening), and the composite score
$\sone$. Local scores are computed on a held-out validation split of
the released training data using the official evaluator, whereas
official challenge test scores represent feedback received exclusively
for submitted checkpoints.

\subsection{Main Result and Local-Hidden Score Analysis}
\label{sec:headline}

\begin{table}[t]
\centering
\caption{Local re-evaluation and official challenge test feedback for the
queried long-schedule checkpoint.}
\label{tab:headline}
\setlength{\tabcolsep}{3pt}
\renewcommand{\arraystretch}{1.05}
\small
\begin{tabular}{l|cccc}
\toprule
Evaluation & $\sloc$ & $\scls$ & $\sscr$ & $\sone$ \\
\midrule
Local official       & 0.826 & 0.436 & 0.543 & 0.596 \\
Challenge test feedback      & 0.489 & 0.465 & 0.696 & 0.564  \\
\midrule
$\Delta$ (local$\to$official test) & $-0.337$ & $+0.029$ & $+0.153$ & $-0.032$ \\
\bottomrule
\end{tabular}
\end{table}

Table~\ref{tab:headline} summarizes the four Track~1 metric components
of the queried long-schedule checkpoint, including the challenge
test feedback and the local re-evaluation results. While the two composite scores are close ($0.596$ v.s. $0.564$), the component-level balance shifts substantially: the local re-evaluation is dominated by localization performance, whereas the official test feedback reveals a stronger image-level screening
signal.

This local-to-official-test discrepancy motivates the adoption of a conservative
experimental protocol. The official test feedback is treated solely as test-time
evidence for the submitted checkpoint and is not used to draw component-level conclusions. Instead, the component-level conclusions are derived from independently trained short-schedule variants and are evaluated
under a unified local protocol.

\subsection{Ablation Study}
\label{sec:abl-modules}

\begin{table}[t]
\centering
\caption{Targeted local ablation under a unified 600-epoch protocol.
The queried 2000-epoch checkpoint is included as a local re-evaluation
reference only.}
\label{tab:abl-modules}
\setlength{\tabcolsep}{3pt}
\renewcommand{\arraystretch}{1.05}
\small
\begin{tabular}{@{}lcccc@{}}
\toprule
Variant & $\sloc$ & $\scls$ & $\sscr$ & $\sone$ \\
\midrule
\multicolumn{5}{@{}l}{\textit{Short schedule (600 ep.)}} \\
\quad Full recipe                       & 0.819 & 0.451 & \textbf{0.629} & \textbf{0.632} \\
\quad no MixRand                        & 0.818 & 0.450 & 0.624 & 0.630 \\
\quad MixRand, no attenuation           & \textbf{0.821} & \textbf{0.453} & 0.622 & 0.631 \\
\quad no SWAD avg.                      & 0.817 & 0.428 & 0.602 & 0.614 \\
\midrule
\multicolumn{5}{@{}l}{\textit{2000-ep queried reference (local re-eval.)}} \\
\quad Full recipe &
  0.826 & 0.436 & 0.543 & 0.596 \\
\bottomrule
\end{tabular}
\end{table}

\begin{figure}[t]
\centering
\includegraphics[width=0.95\columnwidth]{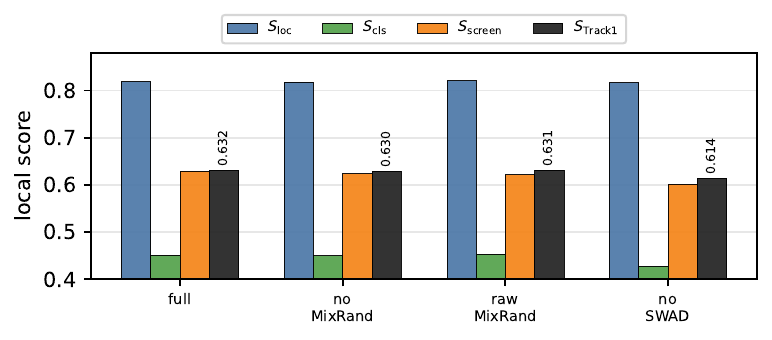}
\caption{Local Track~1 component scores for the comparable 600-epoch
ablations. The no-SWAD variant shows the largest drop on $\sscr$ and $\scls$,
whereas the MixRand variants remain close to the full recipe.}
\label{fig:contrib}
\end{figure}

Table~\ref{tab:abl-modules} presents the targeted local ablation results. While the queried 2000-epoch
checkpoint is also included as a local re-evaluation reference, it does not constitute a valid single-module comparison as it employs a different training budget. Note that its lower local composite score (0.596 vs.\ 0.632 for the 600-epoch full recipe) reflects a different screening-localization balance on the held-out split and should not be interpreted as inferior performance.

Within the comparable 600-epoch ablation set, disabling MixRand only yields a marginal change in the composite score ($0.632$ v.s. $0.630$), and the unattenuated MixRand variant achieves a similar result ($0.631$). These results suggest that attenuation does not degrade performance relative to the raw variant, but they do not demonstrate a large independent performance gain on the current local split. In contrast, removing SWAD checkpoint averaging produces the most pronounced performance drop from $0.632$ to $0.614$, driven primarily by degradation in $\sscr$ and $\scls$. We therefore identify metric-aligned SWAD selection as the strongest measured contributor to overall performance under the current ablation protocol.

Fig.~\ref{fig:contrib} illustrates these results by decomposing
the composite score into the four official metric components. The
no-SWAD variant exhibits the most severe component-level degradation,
further corroborating the importance of selecting and averaging
checkpoints using the Track~1-oriented proxy rather than relying solely
on detector-oriented fitness.

Although the ablation framework supports additional single-variable
variants (e.g., no-IBN, no-SPD, no-SimAM, no-NWD, and no-VFL), the
training budget allocated for this paper does not permit including all
of them in the main table. We therefore treat IBN-a, SPD, SimAM, NWD,
and VFL as fixed recipe components motivated by prior work and by the
small-defect detection and screening-oriented nature of the task,
rather than as independently validated sources of performance gain.

\begin{figure}[t]
\centering
\includegraphics[width=0.95\columnwidth]{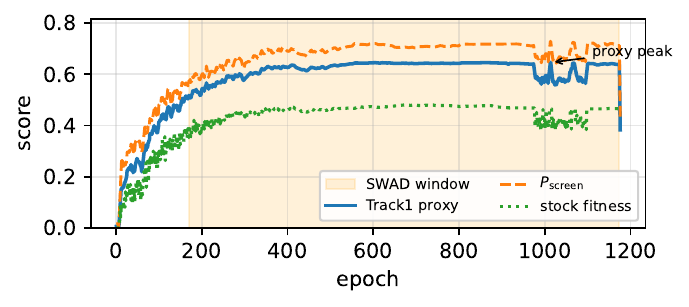}
\caption{Training dynamics of the full recipe.
Solid: smoothed Track~1 proxy $F_{\text{T1}}$ (Eq.~\ref{eq:proxy}).
Dashed: $P_{\text{screen}}$ component alone.
Dotted: default mAP-based detector fitness ($0.1\cdot$mAP50 $+ 0.9\cdot$mAP50-95).
The shaded interval is the SWAD averaging window selected from the
screening-weighted proxy rather than from the default detector fitness.}
\label{fig:curves}
\end{figure}

\begin{figure*}[!t]
\centering
\includegraphics[width=\textwidth,keepaspectratio]{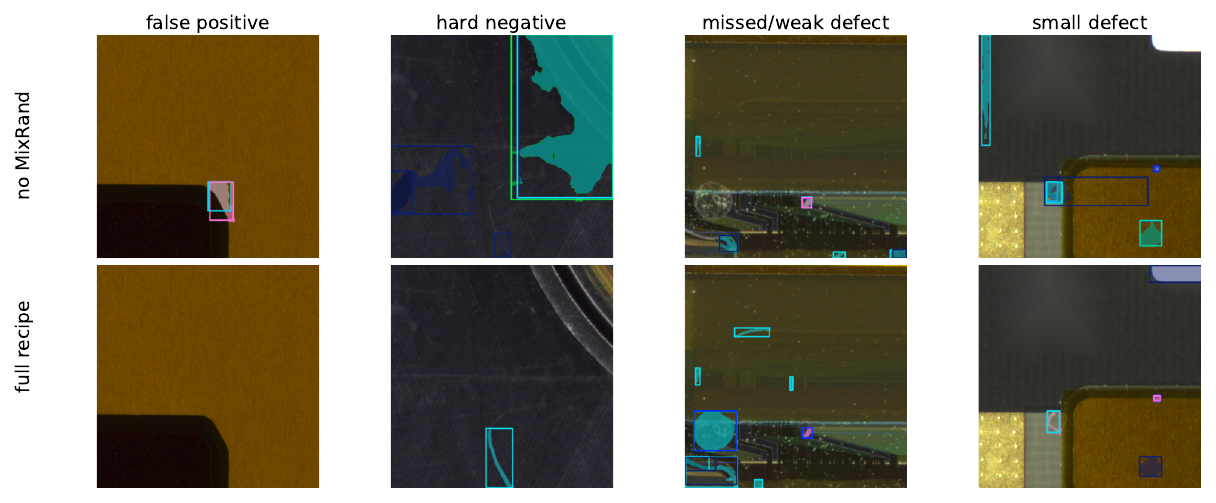}
\caption{Qualitative predictions on the local validation split.
Top row: full recipe with MixRand disabled (the ``no\_mixrand''
ablation). Bottom row: full recipe. Columns show selected clean and
defect-present cases from the local validation split. The figure is
illustrative; quantitative claims are reported in
Table~\ref{tab:abl-modules}.}
\label{fig:qual}
\end{figure*}

\subsection{MixRand Decomposition}
\label{sec:abl-mixrand}

The ``\textit{no MixRand}'' and ``\textit{MixRand, no attenuation}'' rows in Table~\ref{tab:abl-modules} further examine whether composite feature perturbation and unattenuated perturbation strength affect the final performance. Since the observed differences are small on
the current local split, we interpret MixRand primarily as a
stability-oriented regularizer in this setting, consistent with its role
of reducing sensitivity to source-domain appearance statistics rather
than producing a large independent score gain.

The DSU-only and RandConv-only variants are implemented within the ablation
framework but are excluded from the main ablation table due to training
budget constraints. We accordingly rely on prior MixStyle, DSU, and
RandConv studies as motivation for the composite design, rather than
claiming separately verified subcomponent gains on this split.

\subsection{Training-Schedule Effect}
\label{sec:schedule}

Fig.~\ref{fig:curves} compares the evolution of the Track~1 proxy
$\hat{S}_\text{Track1}$, its screening-related component
$P_\text{screen}$, and the default mAP-based detector fitness along the
training trajectory. Their peaks occur at different epochs, showing that
the mAP-oriented fitness is not consistently aligned with the
screening-aware Track~1 objective. Following the flat-minima intuition
of SWAD~\cite{cha2021swad}, our custom driver identifies a flat,
high-proxy plateau using the Track~1-oriented score and averages the
checkpoints within it, rather than anchoring selection to the raw
detector-fitness peak. This divergence motivates the Track~1 proxy as a
more suitable stopping and averaging criterion for this
screening-oriented setting than the default detector fitness.

The no-SWAD control in Table~\ref{tab:abl-modules} provides the
corresponding quantitative evidence: removing SWAD checkpoint averaging
reduces $\sone$ from $0.632$ to $0.614$ and $\sscr$ from $0.629$ to
$0.602$ under the same 600-epoch local protocol.

\subsection{Qualitative Comparison}
\label{sec:qual}

Fig.~\ref{fig:qual} provides a qualitative illustration of the MixRand
ablation by comparing the full recipe with the no-MixRand variant on
selected local-validation images. Cases are chosen to highlight salient
differences, including false positives on clean boards
and missed or weakly detected small defects; they are not random
samples. The figure suggests that the full recipe tends to suppress
spurious detections on defect-free regions while maintaining responses
on small defect instances. Because qualitative examples are inherently
selective, all quantitative performance claims are grounded in the
official component scores reported in Table~\ref{tab:abl-modules}.


\section{Conclusion}
We studied cross-scenario industrial defect detection under domain
shift on the IDA~2026 Track~1 benchmark. Our analysis shows that conventional mAP-driven checkpoint selection can disagree with the deployment objective, especially when robust image-level screening matters as much as defect localization. To address this issue, we presented a domain-robust framework that includes appearance-robust training, small-defect preservation, screening-aware calibration, and metric-aligned checkpoint averaging.
Under a single-model setting with no external data, ensemble, or test-time augmentation, our best queried official challenge test feedback reaches $\sone{=}0.564$ with $\sscr{=}0.696$ on one RTX~4090 GPU. Our results demonstrate that screening is important for this
benchmark and the cross-scenario industrial inspection can benefit from
aligning model selection with deployment-oriented objectives rather than
localization alone. We hope this perspective is useful for future
industrial inspection benchmarks where local validation is an imperfect
proxy for official challenge test performance.

It is worth noting that the targeted local ablation supports the strongest claim for metric-aligned checkpoint selection. The gains from feature perturbation are modest on the current split, and the fixed minor components are
motivated by prior work rather than independently ablated here. Broader official challenge test feedback would be needed to fully separate their effects.

\bibliographystyle{IEEEtran}
\bibliography{references}

\end{document}